\documentclass[11pt,a4paper]{article}

\usepackage{amsmath,amsfonts,amssymb}
\usepackage{graphicx}
\usepackage{algorithm}
\usepackage{algorithmic}
\usepackage{array}
\usepackage{multirow}
\usepackage{subfig}
\usepackage{url}
\usepackage{booktabs}
\usepackage[margin=1in]{geometry}
\usepackage{authblk}
\usepackage{titlesec}

\titleformat{\section}{\normalfont\Large\bfseries}{\thesection.}{0.5em}{}
\titleformat{\subsection}{\normalfont\large\bfseries}{\thesubsection.}{0.5em}{}
\titleformat{\subsubsection}{\normalfont\normalsize\bfseries}{\thesubsubsection.}{0.5em}{}

\newenvironment{IEEEkeywords}{\noindent\textbf{Keywords:} }{\par}

\begin{document}

\title{\Large\bfseries GraPLUS: Graph-based Placement Using Semantics for Image Composition}

\author[1]{Mir Mohammad Khaleghi}
\author[1]{Mehran Safayani\thanks{Corresponding author: M. Safayani (email: safayani@iut.ac.ir).}}
\author[1]{Abdolreza Mirzaei}
\affil[1]{Department of Electrical and Computer Engineering, Isfahan University of Technology, Isfahan 84156-83111, Iran\\
E-mails: m.khaleghi@ec.iut.ac.ir; safayani@iut.ac.ir; mirzaei@iut.ac.ir}

\date{}

\maketitle

\begin{abstract}
We present GraPLUS (Graph-based Placement Using Semantics), a novel framework for plausible object placement in images that leverages scene graphs and large language models. Our approach uniquely combines graph-structured scene representation with semantic understanding to determine contextually appropriate object positions. The framework employs GPT-2 to transform categorical node and edge labels into rich semantic embeddings that capture both definitional characteristics and typical spatial contexts, enabling nuanced understanding of object relationships and placement patterns. GraPLUS achieves placement accuracy of 92.1\% and an FID score of 28.83 on the OPA dataset, outperforming state-of-the-art methods by 8.1\% while maintaining competitive visual quality. In human evaluation studies involving 964 samples assessed by 19 participants, our method was preferred in 52.1\% of cases, significantly outperforming previous approaches (26.3\% and 21.6\% for the next best methods). The framework's key innovations include: (i) leveraging pre-trained scene graph models that transfer knowledge from other domains, eliminating the need to train feature extraction parameters from scratch, (ii) edge-aware graph neural networks that process scene semantics through structured relationships, (iii) a cross-modal attention mechanism that aligns categorical embeddings with enhanced scene features, and (iv) a multiobjective training strategy incorporating semantic consistency constraints. Extensive experiments demonstrate GraPLUS's superior performance in both placement plausibility and spatial precision, with particular strengths in maintaining object proportions and contextual relationships across diverse scene types.
\end{abstract}


\section{Introduction}

Image composition is a fundamental computer vision task that aims to seamlessly integrate multiple images or components into a unified visual scene. A crucial aspect of this process is object placement, determining the optimal position and scale for inserting foreground objects into background scenes. This task has broad applications across creative industries, augmented reality, e-commerce, and data augmentation for computer vision tasks. However, achieving plausible object placement remains challenging as it requires a deep understanding of the semantics of the scene, spatial relationships, and physical constraints.

Early approaches to object placement relied primarily on hand-crafted rules and geometric constraints, which proved limited in their ability to generalize across diverse scenes and object types. Recent learning-based methods have shown promising results, particularly through GAN-based architectures. TERSE~\cite{terse} and PlaceNet~\cite{placenet} pioneered the use of adversarial training to learn placement distributions, while GracoNet~\cite{graconet} introduced graph-based modeling of object relationships. CA-GAN~\cite{ca_gan} proposed a coalescing attention mechanism to enhance placement rationality by capturing self- and cross-feature interactions, while CSANet~\cite{csanet} introduced a convolution scoring approach combined with pyramid grouping to capture multi-scale contextual information efficiently.

Beyond GANs, more recent methods have explored diverse generative models and techniques. DiffPop~\cite{diffpop} utilizes a diffusion probabilistic model for object placement, guided by a structural plausibility classifier to generate placements that are both plausible and diverse. CSENet~\cite{csenet} takes a unique multimodal approach, leveraging large pre-trained language models to incorporate common-sense reasoning and semantic guidance for placement tasks. For a detailed comparison of these and other approaches, please refer to the related work section.

Despite these advances, existing approaches exhibit fundamental limitations in their ability to fully capture the semantic complexity of object placement. GAN-based methods often rely heavily on pixel-level features, missing the structured relational context that governs object arrangements in real scenes. Most critically, current approaches depend on pixel-level analysis of both foreground and background images, increasing computational complexity and limiting generalization. Additionally, these methods typically require end-to-end training of all parameters from scratch, failing to leverage valuable knowledge already encoded in pre-trained models. Graph-based approaches make progress in modeling relationships but lack the semantic richness needed to guide truly contextual placements, while attention-based models struggle with computational efficiency and newer transformer and diffusion-based approaches face resource demands and training instabilities.

To address these limitations, we propose a novel GAN-based framework that fundamentally shifts from pixel-based processing to a structured semantic approach. Two key innovations distinguish our method: First, unlike existing approaches, we determine optimal placement using only the foreground object's category without requiring the actual foreground image, significantly reducing computational complexity while improving generalization across object instances. Second, we leverage transfer learning principles by building upon pre-trained scene graph extraction models that incorporate cross-domain knowledge of common object relationships and spatial arrangements, eliminating the need to train feature extraction parameters from scratch—a significant advantage that enables our model to transfer semantic patterns learned from other domains to the object placement task.

Our method leverages scene graphs as the primary semantic foundation, representing objects and their relationships as nodes and edges within a graph structure that captures the inherent semantic organization of background scenes. This rich structural representation is processed through edge-aware graph neural networks that preserve and enhance semantic relationships between scene elements. While language models play a supporting role, the core of our semantic understanding derives from operations on the scene graph itself, including multi-stage feature enhancement, spatial context integration, and graph transformer networks.

The proposed generator employs a cross-modal attention mechanism that bridges scene understanding and placement generation by using the foreground object's categorical embedding to selectively attend to relevant scene features. The discriminator evaluates placement plausibility through a progressive architecture capturing both fine-grained composition details and global placement coherence. This adversarial approach, combined with our semantic-first design, enables contextually appropriate object placements with improved coherence and accuracy.

Experimental results demonstrate significant improvements in both plausibility of placement and spatial precision. The proposed approach outperforms existing methods in center accuracy and object scale preservation, achieving the lowest mean center distance to ground truth and the highest proportion of placements with optimal scale ratios. Human evaluation confirms these findings, with our user study (964 samples, 19 participants) showing that placements generated by our method were preferred in 52.1\% of cases, substantially outperforming GracoNet (26.3\%) and CSANet (21.6\%). These enhancements enable contextually appropriate, semantically coherent, and geometrically precise placements across various scene configurations.

The main contributions of our work can be summarized as follows:
\begin{itemize}
  \item{We introduce a novel GAN-based framework that leverages semantic scene graphs and language models for object placement, using only the foreground object's category rather than its image and prioritizing semantic plausibility over raw diversity.}
  \item{We develop an innovative architecture combining graph neural networks with attention mechanisms that builds on transfer learning from pre-trained scene graph extraction models, eliminating the need to train feature extraction parameters from scratch.}
  \item{Through extensive experiments, we demonstrate significant improvements over existing approaches in placement plausibility, validating the effectiveness of our semantic-first approach in generating contextually appropriate object placements.}
\end{itemize}

\section{Related Work}

Object placement methodologies have evolved across three distinct paradigms: rule-based approaches, learning-based generative methods, and hybrid architectures.

Early research relied on deterministic systems using geometric constraints and heuristics to position foreground objects within scenes~\cite{trad1,trad2,trad3,trad4,trad5}. These methods laid foundational principles by addressing specific aspects like depth preservation and contextual similarity. However, they struggled to generalize due to limited contextual awareness and often relied on overly simplistic assumptions about spatial relationships.

The emergence of deep learning introduced generative approaches, with GANs becoming a cornerstone in this field. TERSE~\cite{terse} demonstrated adversarial learning for synthetic data generation, but suffered from limited diversity and rigid placement distributions. PlaceNet~\cite{placenet} advanced this framework by incorporating inpainting mechanisms to enhance placement distribution learning. However, it struggled to maintain semantic coherence in complex scenes. GracoNet~\cite{graconet} made a significant shift by redefining the problem as graph completion, introducing structured scene representation through graph neural networks. Despite this innovation, its graph-based approach often failed to capture fine-grained spatial relationships and lacked semantic richness needed for truly contextual placements.

Attention mechanisms emerged as a key innovation in subsequent approaches, addressing some of the limitations of earlier methods. CA-GAN~\cite{ca_gan} pioneered the use of coalescing attention for modeling interactions between foreground and background elements, significantly improving placement plausibility. However, its computational complexity limited applicability to high-resolution images. CSANet~\cite{csanet} further developed this direction through multi-scale feature interaction and pyramid pooling, achieving better computational efficiency. Nevertheless, it still struggled with preserving object proportions in varied scene contexts and lacked a semantic understanding of object relationships.

The field has recently expanded beyond pure GAN architectures to explore diverse methodologies, each with their own strengths and weaknesses. Transformer-based approaches like TopNet~\cite{topnet} introduced heatmap generation for placement prediction, but often lacked precision in scale estimation and required substantial computational resources. Reinforcement learning methods such as IOPRE~\cite{iopre} enabled iterative refinement through interactive feedback, yet suffered from unstable training dynamics and complex optimization procedures. Diffusion-based models like DiffPop~\cite{diffpop} showed promise in balancing diversity and plausibility, though at the cost of increased inference time and training complexity. Multimodal integration has also gained prominence, with models like CSENet~\cite{csenet} incorporating cross-modal understanding, but these approaches often require large-scale pre-trained models with substantial computational demands and fail to fully leverage the structured nature of scene composition.

In contrast to these methods, our approach addresses these limitations by integrating scene graph structures with semantic embeddings, enabling a more comprehensive understanding of object-scene relationships while maintaining computational efficiency through transfer learning from pre-trained models.

\section{Proposed Method}

\subsection{Overview}
Object placement in computer vision requires precise positioning of foreground objects within background scenes while maintaining semantic and visual coherence. Our framework addresses this fundamental challenge through a novel semantic-first approach that achieves state-of-the-art performance on the Object Placement Assessment (OPA) dataset~\cite{opa} through several key architectural innovations.

\begin{figure}
    \centering
    \includegraphics[width=\textwidth]{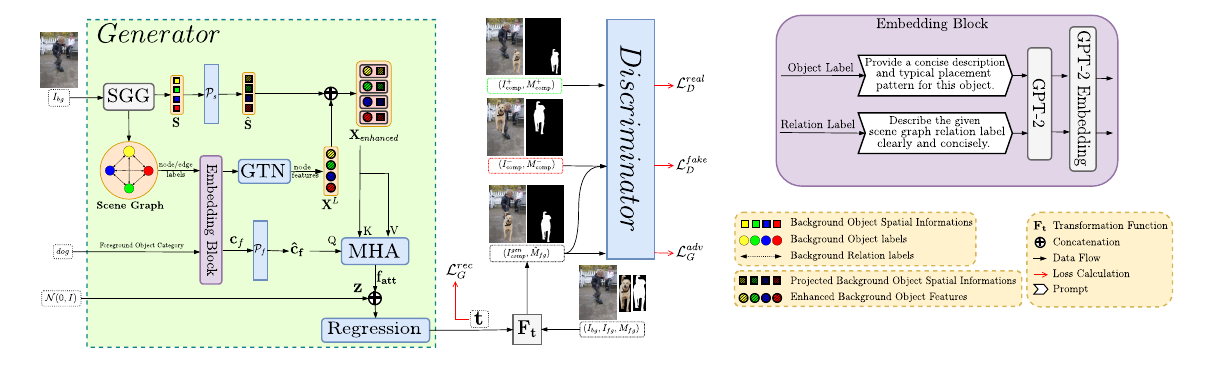}
    \caption{Overview of the GraPLUS framework architecture. The generator (left, green) processes background images through scene graph generation and enhancement, while the discriminator (center) evaluates composition quality. The embedding block (right, purple) provides semantic enrichment using GPT-2. The legend (bottom right) explains the visual elements used in the diagram.}
    \label{fig:framework}
\end{figure}

As illustrated in Fig.~\ref{fig:framework}, the framework comprises four principal components. The scene graph processing module transforms an input background image $I_{\text{bg}} \in \mathbb{R}^{3 \times H \times W}$ into a structured graph representation, capturing object instances and their inter-relationships. Subsequently, a two-stage enhancement process first maps nodes and edges to learnable embeddings and then augments these representations with spatial information derived from object bounding boxes, significantly improving the model's placement accuracy over the base architecture.

The enhanced representations undergo processing through our Graph Transformer Network (GTN), which employs edge-aware attention to model object-object interactions. Given a foreground object category $c_f$, a cross-attention module computes the attention weights between the projected embedding of the object $\hat{c}_f$ and the graph features to identify relevant regions of the scene. These attention-weighted characteristics guide the placement network in predicting transformation parameters $t = [t_r, t_x, t_y] \in \mathbb{R}^3$, where $t_r$ controls the scale and $(t_x, t_y)$ determine the translation coordinates. The integration of cross-attention mechanisms further enhances the model's ability to generate plausible object placements.

In the following sections, we provide a detailed explanation of each component within our framework. We begin with the scene graph generation process, outlining how structured representations are derived from input images and enriched with GPT-based semantic embeddings. Next, we describe the scene graph enhancement stage, where the Graph Transformer Network (GTN) refines spatial and relational features to improve object placement accuracy. We then introduce the cross-modal attention mechanism, which aligns the foreground object representation with scene features to determine optimal placement. Following this, we discuss transformation prediction and image composition, detailing how placement parameters are estimated and applied to generate composite images. Finally, we elaborate on the adversarial training methodology, covering the optimization process, loss functions, and the overall strategy to refine object placements.

\subsection{Scene Graph Generation}

\begin{figure}
    \centering
    \includegraphics[width=0.8\textwidth]{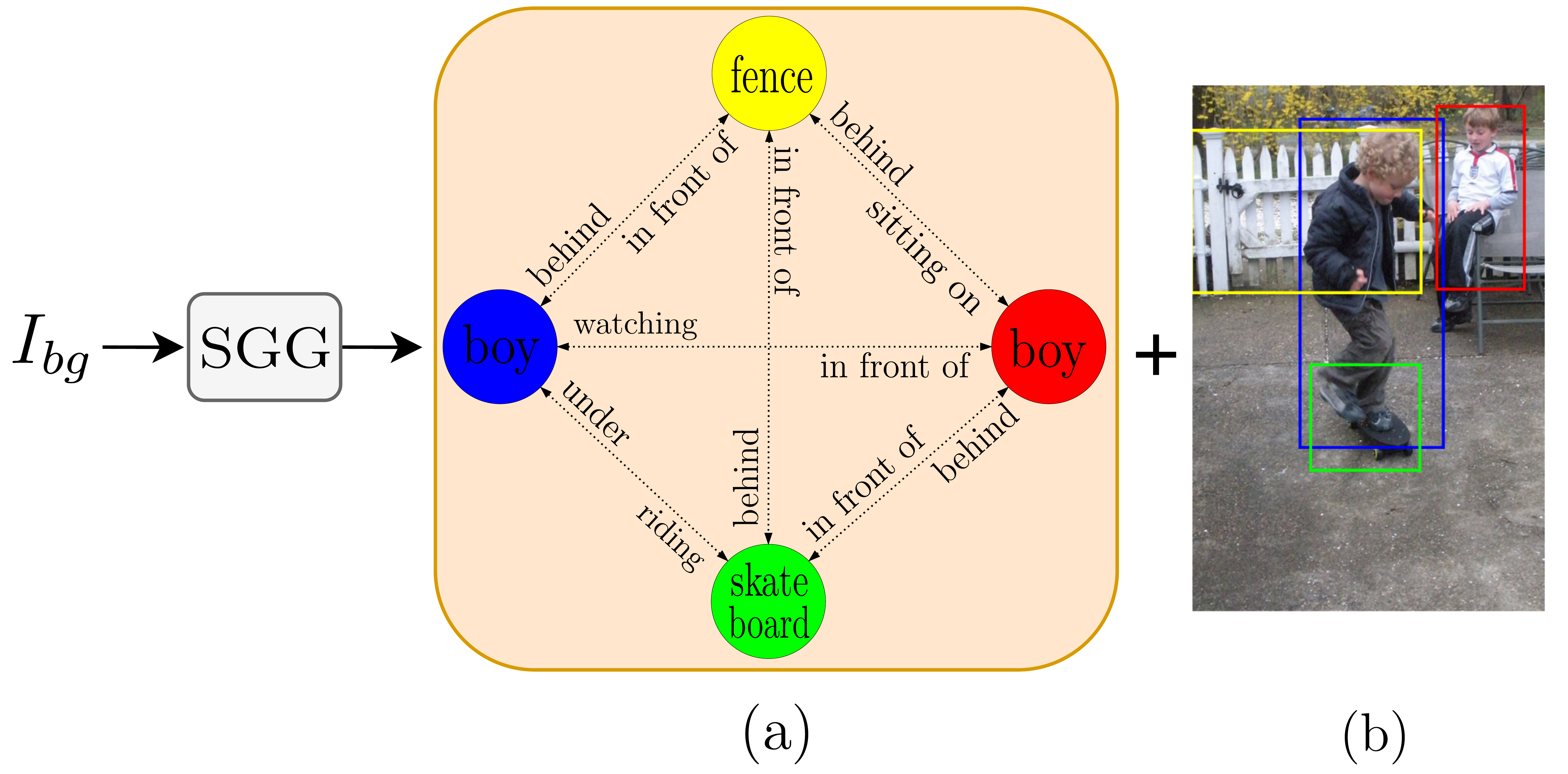}
    \caption{Overview of Scene Graph Generation (SGG). Given an input image ($I_{bg}$), the SGG module generates (a) a scene graph capturing object relationships and their interactions (nodes represent objects and edges represent relationships), and (b) corresponding bounding box detections for each object.}
    \label{fig:sgg}
\end{figure}

The foundation of our framework lies in the generation and semantic enhancement of scene graphs, which provide a structured representation of visual scenes. This section outlines our methodology for constructing and enriching these graph representations to support downstream object placement reasoning.

Given an input image $I_{\text{bg}}$, we employ the Neural-MOTIFS architecture~\cite{neural_motifs} for scene graph generation. As illustrated in Fig.~\ref{fig:sgg}, the SGG module transforms the input image $I_{bg}$ into a structured scene graph representation along with corresponding bounding box detections. The scene graph captures both spatial and interactive relationships between objects, while the bounding boxes provide geometric anchoring, enabling our framework to reason about both semantic relationships and spatial constraints during object placement.

The scene graph is represented as a tuple $(O, E)$, where $O = \{o_1, \ldots, o_n\}$ denotes the set of detected objects, with each $o_i \in C_{\text{obj}}$ derived from the object category vocabulary $C_{\text{obj}}$ of 151 categories. The set $E \subseteq O \times R \times O$ represents directed edges of the form $(o_i, r_{ij}, o_j)$, where $o_i, o_j \in O$ and $r_{ij} \in R$, with $R$ indicating the set of possible relationships drawn from a relationship category vocabulary $C_{\text{rel}}$ of 51 relation types. These relationships span various categories including geometric relations (e.g., ``above,'' ``behind,'' ``under''), possessive relations (e.g., ``has,'' ``part of,'' ``wearing''), and semantic relations (e.g., ``carrying,'' ``eating,'' ``using''). For example, in a typical scene, a person might be ``wearing'' clothes, ``riding'' a vehicle, or ``holding'' an object, while furniture items might be related through geometric relationships like ``on'' or ``next to.''

Since the generated scene graph is a complete graph, each object is connected to every other object in the scene. Consequently, the number of edges is given by $m = 2 \times \binom{n}{2} = n(n-1)$, where $n$ is the number of detected objects. This fully connected structure allows the model to account for all possible pairwise relationships, ensuring a comprehensive understanding of spatial and semantic interactions.

For each detected object $o_i$, we extract its corresponding bounding box coordinates $b_i \in \mathbb{R}^4$, forming the set $B = \{b_1, \ldots, b_n\}$ of all bounding boxes. Each bounding box $b_i$ encodes the spatial information $(x, y, w, h)$, representing the center coordinates, width, and height of the detected object in the image coordinate space.

Scene graph generation is performed using the pre-trained Neural-MOTIFS model \cite{neural_motifs} operating in Total Direct Effect (TDE) mode~\cite{TDE}. The model was originally trained on the Visual Genome dataset, which provides a rich collection of annotated images with objects, attributes, and relationships. This approach mitigates dataset biases by effectively separating causal relationships from contextual correlations. The pre-trained model has learned to identify recurring patterns (motifs) in scene structures, allowing it to transfer knowledge of common object relationships and spatial arrangements across different domains. This cross-domain knowledge transfer represents a significant advantage of our approach, as it enables our placement framework to leverage semantically rich representations that capture both common sense knowledge and domain-specific visual patterns. The resulting structured representation encapsulates both spatial information via bounding box coordinates and semantic relationships through predicted edges, forming a robust foundation for the object placement framework.

\subsection{Semantic Feature Enrichment via Language Models}

The initial scene graph representation is enhanced using semantic embeddings derived from GPT-2~\cite{radford2019language}, which transform categorical node and edge labels into dense vector representations that capture semantic relationships and contextual information. The descriptive dictionaries for the categories of objects and relationships, including definitional characteristics and typical spatial contexts, were automatically generated by GPT, ensuring consistency and comprehensive coverage.

For each node of the object $o_i \in O$, its categorical label is assigned to a semantic embedding vector $\hat{o}_i \in \mathbb{R}^{768}$ through the GPT-2 encoder. Similarly, each relationship label $r_{ij} \in R$ is assigned to an embedding $\hat{r}_{ij} \in \mathbb{R}^{768}$. These embeddings are generated by processing descriptive templates that integrate object definitions and spatial placement contexts.

For example, the category \textit{bicycle} is associated with the description \textit{``pedal vehicle''} and the placement context \textit{``on paths, properly balanced.''} Similarly, the category \textit{tree} is described as a \textit{``tall plant''} with placement specified as \textit{``in ground, natural growth.''} These detailed descriptions enable GPT-2 to generate embeddings that encapsulate nuanced semantic relationships and spatial patterns.

The node embedding process incorporates the 151 object categories supported by the scene graph generator and the 47 categories from the OPA dataset, ensuring extensive coverage of potential object types. Each node category is processed through multiple semantic perspectives, including its definition, spatial placement patterns, and contextual relationships, generating rich semantic representations.

When processing a foreground object for placement, its category is embedded similarly to obtain ${c}_f \in \mathbb{R}^{768}$, facilitating direct semantic comparison with the elements of the scene graph. The embedding process for scene graph components and foreground objects leverages GPT-2's pretrained knowledge to effectively capture semantic relationships and typical spatial arrangements, significantly enhancing the quality of the scene graph representation.

\subsection{Graph Enhancement}

The Graph Transformer Enhancement module is a key component of our framework, designed to process and enrich the semantically embedded scene graph representation through a dual-stage approach. Initially, a graph transformer network (GTN) refines node and edge features while preserving their structural relationships using an edge-aware multihead attention mechanism. Subsequently, this representation is augmented with explicit spatial information through a dedicated spatial projection layer, enabling the network to simultaneously reason about semantic and geometric constraints. This two-stage enhancement allows our model to capture both semantic relationships between scene elements and their precise spatial configurations, significantly improving placement accuracy.

\subsubsection{Graph Transformer Network}

After the initial semantic embedding phase, the representation of the scene graph is refined through the Graph Transformer Network (GTN) architecture, which processes node and edge features concurrently while maintaining their structural relationships. The GTN employs an edge-aware multi-head attention mechanism with $H_G$ attention heads per layer to model complex interdependencies within the scene graph structure.

For each layer, let $d_{in}$ and $d_{out}$ denote the input and output dimensions, respectively, with $d_{in} = 768$ corresponding to the embedding dimension of GPT-2 for the first layer. The nodes of the characteristic matrix on the layer $l-1$ are represented as $\mathbf{X}^{l-1} \in \mathbb{R}^{n \times d_{in}}$, where $n$ is the number of nodes. Similarly, the edge feature matrix for $m$ edges is given by $\mathbf{E}^{l-1} \in \mathbb{R}^{m \times d_{in}}$. The adjacency structure is denoted by $\mathcal{A} \in \{0,1\}^{n \times n}$, where $\mathcal{A}_{ij} = 1$ indicates an edge between the nodes $i$ and $j$.

The transformation process at each layer is expressed as:
\begin{equation}
\mathbf{Q}^l = \mathbf{X}^{l-1}\mathbf{W}_Q^l,\; \mathbf{K}^l = \mathbf{X}^{l-1}\mathbf{W}_K^l,\; \mathbf{V}^l = \mathbf{X}^{l-1}\mathbf{W}_V^l
\end{equation}
where $\mathbf{W}_Q^l, \mathbf{W}_K^l, \mathbf{W}_V^l \in \mathbb{R}^{d_{in} \times d_h}$ are learnable projection matrices, and $d_h = d_{out} / H_G$ denotes the dimensionality per attention head. Edge features are projected similarly:
\begin{equation}
\mathbf{E}^l = \mathbf{E}^{l-1}\mathbf{W}_E^l
\end{equation}
where $\mathbf{W}_E^l \in \mathbb{R}^{d_{in} \times d_h}$ is the learnable weight matrix for edge features.

Edge-aware attention scores for node pairs $(i, j)$ are calculated as:
\begin{equation}
\alpha_{ij}^{edge,l} = \frac{(\mathbf{q}_i^l)^\top \mathbf{k}_j^l + (\mathbf{q}_i^l)^\top \mathbf{e}_{ij}^l + (\mathbf{e}_{ij}^l)^\top \mathbf{k}_j^l}{\sqrt{d_h}}
\end{equation}
where $\mathbf{q}_i^l, \mathbf{k}_j^l, \mathbf{v}_j^l \in \mathbb{R}^{d_h}$ are rows from $\mathbf{Q}^l, \mathbf{K}^l, \mathbf{V}^l$, and $\mathbf{e}_{ij}^l \in \mathbb{R}^{d_h}$ represents edge features. Self-attention scores are computed as:
\begin{equation}
\alpha_{ii}^{self,l} = \frac{(\mathbf{q}_i^l)^\top \mathbf{k}_i^l}{\sqrt{d_h}}
\end{equation}

The attention score matrix $\mathbf{A}^l \in \mathbb{R}^{n \times n \times H_G}$ integrates edge-aware and self-attention scores:
\begin{equation}
\mathbf{A}_{ij}^l = \begin{cases}
\alpha_{ij}^{edge,l} & \text{if } i \neq j \text{ and } \mathcal{A}_{ij} = 1 \\
\alpha_{ii}^{self,l} & \text{if } i = j \\
-\infty & \text{otherwise}
\end{cases}
\end{equation}

Normalized attention weights are obtained via softmax normalization:
\begin{equation}
\hat{\alpha}_{ij}^l = \text{softmax}(\mathbf{A}_{ij}^l)
\end{equation}

Node features are updated using a message-passing mechanism:
\begin{equation}
\mathbf{X}^l = \text{LayerNorm}(\mathbf{X}^{l-1} + \text{MLP}(\sum_{j=1}^n \hat{\alpha}_{ij}^l \mathbf{v}_j^l))
\end{equation}
where $\text{MLP}(\cdot)$ denotes a multi-layer perceptron, and $\text{LayerNorm}(\cdot)$ applies layer normalization. The final node features $\mathbf{X}^L \in \mathbb{R}^{n \times d_{out}}$ encapsulate comprehensive scene information for subsequent tasks, such as placing objects and making decisions.

The implementation of this architecture addresses computational efficiency and memory management concerns. Our approach optimizes batch processing by handling node features in batches of size $B$, with attention computations executed in parallel across all nodes within each batch. Edge features are stored using an efficient sparse representation to minimize memory requirements, while attention weights are computed using masked softmax operations to handle varying graph sizes within the same batch efficiently.

\subsubsection{Spatial Information Integration}

The output of the GTN, $\mathbf{X}^L \in \mathbb{R}^{n \times d_{out}}$, is enhanced by explicit spatial information for each node. A spatial feature vector $\mathbf{s}_i \in \mathbb{R}^9$ encodes geometric information:
\begin{equation}
\mathbf{s}_i = [W_{bg}, H_{bg}, x_i, y_i, w_i, h_i, t_{r_i}, t_{x_i}, t_{y_i}]
\end{equation}
where $(W_{bg}, H_{bg})$ represent background dimensions, and $(x_i, y_i, w_i, h_i)$ denote bounding box parameters. Transformation parameters $(t_{r_i}, t_{x_i}, t_{y_i})$ are defined by aspect ratios and spatial configurations. Specifically, $t_{r_i}$ is calculated as $t_{r_i} = h_i / H_{bg}$ if $AR_{obj_i} < AR_{bg}$, and $t_{r_i} = w_i / W_{bg}$ otherwise. Relative positions are computed as $t_{x_i} = x_i / (W_{bg} - w_i)$ and $t_{y_i} = y_i / (H_{bg} - h_i)$, ensuring that all parameters are within $(0,1)$.

Spatial features are aggregated into $\mathbf{S} \in \mathbb{R}^{n \times 9}$ and projected to a higher dimension:
\begin{equation}
\hat{\mathbf{S}} = \mathcal{P}_s(\mathbf{S})
\end{equation}
where $\mathcal{P}_s: \mathbb{R}^{9} \rightarrow \mathbb{R}^{d_{spatial}}$ is a learned projection function implemented as a linear layer.
The enhanced representation $\mathbf{X}_{enhanced} \in \mathbb{R}^{n \times d_{model}}$ is obtained by concatenating semantic features from the GTN with transformed spatial features, where $d_{model} = d_{out} + d_{spatial}$.

\begin{equation}
\mathbf{X}_{enhanced} = [\mathbf{X}^L \parallel \hat{\mathbf{S}}]
\end{equation}
This combined representation integrates semantic and spatial features, enabling superior object placement and decision-making performance.

\subsection{Cross-Modal Attention Mechanism}

The enhanced scene graph representation $\mathbf{X}_{enhanced}$ must be effectively combined with information about the foreground object to guide the placement decision. We achieve this through a dedicated cross-modal attention mechanism that enables the model to focus on relevant scene regions based on the foreground object category.

The foreground category embedding ${c}_f \in \mathbb{R}^{768}$ is first projected to match the dimensionality of the enhanced scene features:
\begin{equation}
\mathbf{\hat{c}_f}= \mathcal{P}_f({c}_f) \in \mathbb{R}^{d_{model}}
\end{equation}
where $\mathcal{P}_f: \mathbb{R}^{768} \rightarrow \mathbb{R}^{d_{model}}$ is a learned projection function implemented as a linear layer.

The projected category embedding serves as the query in our multi-head attention mechanism, while the enhanced scene features $\mathbf{X}_{enhanced} \in \mathbb{R}^{n \times d_{model}}$ provide the keys and values. The initial projections are computed as:
\begin{equation}
\mathbf{Q} = \mathbf{c}_f\mathbf{W}_Q,\; \mathbf{K} = \mathbf{X}_{enhanced}\mathbf{W}_K,\; \mathbf{V} = \mathbf{X}_{enhanced}\mathbf{W}_V
\end{equation}
where $\mathbf{W}_Q, \mathbf{W}_K \in \mathbb{R}^{d_{model} \times (d_k \cdot H)}$ and $\mathbf{W}_V \in \mathbb{R}^{d_{model} \times (d_v \cdot H)}$ are learnable projection matrices, with $d_k$ and $d_v$ denoting the key and value dimensions per attention head, and $H$ is the number of heads.

To incorporate positional information, we learn position-specific key and value embeddings:
\begin{equation}
\mathbf{K}_{pos} = \text{PE}_{K}(1,\ldots,n) \in \mathbb{R}^{H \times n \times d_k}
\end{equation}
\begin{equation}
\mathbf{V}_{pos} = \text{PE}_{V}(1,\ldots,n) \in \mathbb{R}^{H \times n \times d_v}
\end{equation}
where $\text{PE}_K$ and $\text{PE}_V$ are learned positional embedding functions.

For each attention head $h \in \{1,\ldots,H\}$, the attention weights and outputs are computed as:
\begin{equation}
\mathbf{a}_h = \text{softmax}\left(\frac{\mathbf{Q}_h(\mathbf{K}_h + \mathbf{K}_{pos,h})^\top}{\sqrt{d_k}}\right) \in \mathbb{R}^{1 \times n}
\end{equation}
\begin{equation}
\mathbf{f}_h = \mathbf{a}_h(\mathbf{V}_h + \mathbf{V}_{pos,h}) \in \mathbb{R}^{1 \times d_v}
\end{equation}
where $\mathbf{Q}_h \in \mathbb{R}^{1 \times d_k}$, $\mathbf{K}_h \in \mathbb{R}^{n \times d_k}$, and $\mathbf{V}_h \in \mathbb{R}^{n \times d_v}$ represent the $h$-th head's query, key, and value matrices respectively, obtained by reshaping the initial projections. $\mathbf{K}_{pos,h}$ and $\mathbf{V}_{pos,h}$ are the corresponding positional embeddings for the $h$-th head.

The final attention output is obtained by concatenating the outputs from all heads, applying a final projection, and adding a residual connection:
\begin{equation}
\mathbf{f_{att}} = \text{LayerNorm}([\mathbf{f}_1 \parallel \ldots \parallel \mathbf{f}_H]\mathbf{W}_O + \mathbf{c}_f)
\end{equation}
where $\mathbf{W}_O \in \mathbb{R}^{(d_v \cdot H) \times d_{model}}$ is a learnable projection matrix, $\parallel$ denotes concatenation, and LayerNorm represents layer normalization. The residual connection with $\mathbf{c}_f$ helps maintain the category information throughout the attention process.

This cross-modal attention mechanism effectively conditions the understanding of the scene in the category of objects in the background, producing a focused representation $\mathbf{f_{att}} \in \mathbb{R}^{1 \times d_{model}}$ that captures the most relevant scenes' regions to place the given object. The attention weights $\mathbf{a}_h$ provide interpretable importance scores for each scene node, providing information on the placement decisions of the model.

\subsection{Transformation Prediction and Image Composition}

The final stages of our placement network predict the transformation parameters and generate the composite image using differentiable geometric transformations, following the approach proposed by Zhang et al.~\cite{graconet}. This process, which we denote as function $F_t$, takes the predicted transformation parameters $\mathbf{t}$, background image $I_{bg}$, foreground image $I_{fg}$, and foreground mask $M_{fg}$ as inputs to produce the final composite image $I_{comp}$. This process combines the learned spatial parameters with precise geometric operations for seamless object integration.

\subsubsection{Parameter Regression Network}

The transformation prediction leverages attention-weighted features $\mathbf{f_{att}}$, which are concatenated with a random noise vector $\mathbf{z} \sim \mathcal{N}(0, I)$ to introduce controlled variability in the placement of objects:

\begin{equation}
\mathbf{f_{combined}} = [\mathbf{f_{att}} \parallel \mathbf{z}] \in \mathbb{R}^{1 \times (d_{att} + d_{noise})}
\end{equation}

where $d_{att}$ represents the feature dimension and $d_{noise}$ denotes the noise vector dimension. A multilayer regression network $\mathcal{R}$ processes this combined feature vector. The transformation parameters $\mathbf{t} = [t_r, t_x, t_y]$ are obtained through:

\begin{equation}
\mathbf{t} = 0.5 \tanh(\mathcal{R}(\mathbf{f_{combined}})) + 0.5
\end{equation}

This approach maps the network output to the $[0, 1]$ interval, where $t_r$ controls the scaling, and $(t_x, t_y)$ determine the translation coordinates.

\subsubsection{Differentiable Geometric Transformation}

As part of the function $F_t$, and inspired by the Spatial Transformer Networks (STN) approach~\cite{stn}, the predicted transformation parameters construct an affine transformation matrix $\boldsymbol{\Theta}$:

\begin{equation}
\boldsymbol{\Theta} = \begin{bmatrix}
\frac{1}{t_r + \epsilon} & 0 & (1 - 2t_x)\left(\frac{1}{t_r + \epsilon} - \frac{b_w}{W}\right) \\
0 & \frac{1}{t_r + \epsilon} & (1 - 2t_y)\left(\frac{1}{t_r + \epsilon} - \frac{b_h}{H}\right)
\end{bmatrix}
\end{equation}

where $\epsilon$ ensures numerical stability, $(b_w, b_h)$ represent normalized bounding box dimensions, and $(W, H)$ are image dimensions. The transformation generates a sampling grid $\mathcal{G}$, enabling spatial transformation of the foreground object.

Within $F_t$, the transformed foreground image $\hat{I}_{fg}$ and mask $\hat{M}_{fg}$ are obtained through differentiable bilinear sampling, where $\mathcal{S}$ represents the grid sampling function that interpolates pixel values based on the generated sampling grid $\mathcal{G}$:

\begin{equation}
\hat{I}_{fg} = \mathcal{S}(I_{fg}, \mathcal{G}), \quad \hat{M}_{fg} = \mathcal{S}(M_{fg}, \mathcal{G})
\end{equation}

The final composite image and transformed mask $(I_{comp}^{gen}, \hat{M}_{fg}) = F_t(I_{bg}, I_{fg}, M_{fg}, \mathbf{t})$ are generated by alpha blending, where $\odot$ denotes element-wise multiplication (Hadamard product):

\begin{equation}
I_{comp}^{gen} = \hat{M}_{fg} \odot \hat{I}_{fg} + (1 - \hat{M}_{fg}) \odot I_{bg}
\end{equation}

This operation combines the transformed foreground image $\hat{I}_{fg}$ with the background image $I_{bg}$ using the transformed foreground mask $\hat{M}_{fg}$ as the alpha blending weight. The element-wise multiplication ensures that each pixel of the foreground image and background image is scaled by the corresponding mask value, enabling seamless integration of the foreground object into the background scene.

\subsection{Adversarial Training Methodology}

Our training approach addresses key challenges in object placement and image composition, focusing on stable optimization and effective learning dynamics. The core of our methodology is a carefully designed multi-stage training process that balances generator and discriminator updates.

To assess placement plausibility, we incorporate a discriminator that distinguishes between realistic and implausible placements by processing both the composite image $I_{comp}$ and its corresponding foreground mask $\hat{M}_{fg}$. The discriminator follows a progressive architecture, extracting both fine-grained composition details and global placement coherence. During training, the generator learns to produce plausible object placements by minimizing adversarial loss against this discriminator.

The training process employs a sophisticated optimization strategy within each iteration, drawing inspiration from adversarial training techniques. The objective is to simultaneously optimize the generator and the discriminator while maintaining the stability of the training.

1) The discriminator is first updated on real samples to establish a robust foundation for image recognition:
\begin{equation}
\mathcal{L}_{D}^{real} = \mathbb{E}_{x \sim p^{real}_{data}}[\log D(x)]
\end{equation}
where $p^{real}_{data}$ represents the distribution of real composite images from the dataset.

2) The discriminator is then updated on synthetic samples:
\begin{equation}
\mathcal{L}_{D}^{fake} = \mathbb{E}_{x \sim p^{fake}_{data}}[\log(1 - D(x))] + \mathbb{E}_{x \sim p_{G}}[\log(1 - D(G(x)))]
\end{equation}
where $p^{fake}_{data}$ represents the distribution of fake composite images from the dataset, and $p_G$ represents the distribution of images generated by our model.

3) Finally, the generator is updated using a combined loss:
\begin{equation}
\mathcal{L}_{G}^{total} = \mathcal{L}_{G}^{adv} + \lambda_{rec}\mathcal{L}_{G}^{rec}
\end{equation}
where $\mathcal{L}_{G}^{adv} = \mathbb{E}_{x \sim p_G}[\log D(G(x))]$ is the adversarial loss and $\lambda_{rec}=50$ weights the reconstruction loss.

Following \cite{graconet}, the reconstruction loss for transformation parameters incorporates geometric constraints:
\begin{equation}
\mathcal{L}_{G}^{rec} = \|\mathbf{t} - \mathbf{t}_{gt}\|_2 \odot \mathbf{w}(\mathbf{t})
\end{equation}
where $\|\cdot\|_2$ denotes the L2 norm, $\mathbf{t}$ represents the predicted transformation parameters, $\mathbf{t}_{gt}$ the ground truth parameters, and $\mathbf{w}(\mathbf{t}) = [\sin(t_r\pi/2), \cos(t_r\pi/2), \cos(t_r\pi/2)]$ is an adaptive weight vector.

This meticulously designed training strategy ensures stable optimization while promoting diversity in generated placements. The multistage approach allows for nuanced learning dynamics, enabling the model to capture complex spatial relationships and generate high-quality composite images.

\section{Experiments}

\subsection{Experimental Setup}

\subsubsection{Dataset}
The OPA data set \cite{opa}, a comprehensive benchmark for plausibility of object placement, is used to evaluate our approach. The data set comprises 62,074 (11,396) train (test) images in 47 COCO-derived object categories. The training set includes 21,376 positive (40,698 negative) samples, while the test set contains 3,588 positive (7,808 negative) samples. Rigorous evaluation is ensured by strictly separating background and foreground elements between the training and test sets, minimizing bias and cross-contamination.

\subsubsection{Implementation Details}
Our framework was implemented using PyTorch and executed on an NVIDIA RTX 1080Ti GPU with 12GB VRAM. The optimization process used the Adam algorithm with carefully calibrated learning rates: $1 \times 10^{-4}$ for the generator and $2 \times 10^{-5}$ for the discriminator, utilizing momentum parameters $\beta_1 = 0.5$ and $\beta_2 = 0.999$. Training was carried out with a batch size of $B=32$, ensuring stable optimization while maintaining computational efficiency.

The OPA dataset's inherent class imbalance, with a fake-to-real ratio of ~2:1, necessitated a robust sampling and augmentation strategy. To address this, we developed a dynamic sampling mechanism that balances real and fake samples within training batches. Specifically, real samples that appear multiple times within batches during one epoch are rotated in subsequent epochs, ensuring diversity and mitigating bias and overfitting.

Our data augmentation protocol incorporates probabilistic transformations, including horizontal flipping (applied with a probability of 0.5), color jittering (0.5), Gaussian blurring (0.5), and grayscale conversion (0.2). These augmentations expand the training dataset, introduce controlled variability, and enhance the model's robustness and generalization to visual perturbations.

The discriminator training process was meticulously engineered to address the challenges posed by the imbalance of the data set. By updating the discriminator twice on real samples before processing generated and fake samples, we developed a more nuanced approach to distinguishing placement plausibility, mitigating the potential bias introduced by sample distribution disparities.

Spatial feature processing used a 256-dimensional feature space, striking an optimal balance between capturing fine-grained positional details and maintaining computational efficiency. The Graph Transformer Network was configured with 5 layers and 8 attention heads per layer, enabling sophisticated feature extraction and representation learning. The multihead cross-attention mechanism employed eight attention heads with key and value dimensions of 64, incorporating a residual connection strategy to enhance feature integration and capture complex inter-node relationships. A high-dimensional noise vector of 2048 elements was introduced during transformation prediction, providing substantial stochastic variability to the placement generation process.

\subsection{Evaluation Methodology}
We evaluate our approach using metrics that span three categories: placement quality, generation diversity, and spatial precision.

Placement quality is assessed using three metrics: (1) placement plausibility using an extended SimOPA model~\cite{opa} as a binary classifier, measuring the proportion of generated compositions classified as reasonable; (2) Fréchet Inception Distance (FID)~\cite{fid} between generated and ground truth positive samples, quantifying visual quality across the 3,588 positive samples in the test set; and (3) user study, where the subjective quality of generated compositions is evaluated by human participants. 

For the first metric, we utilize the SimOPA classifier to distinguish between reasonable and unreasonable object placements, with accuracy defined as the proportion of composite images generated classified as reasonable during inference. For the second metric, FID is calculated between our generated composite images and the ground-truth positive samples from the OPA test set, providing a robust measure of visual quality that correlates well with human perception. For the third metric, we conducted a user study selecting 964 unique foreground-background pairs from the positive samples in the test set. These pairs were evenly distributed among 19 voluntary participants, ensuring that there was no overlap between the samples evaluated by each participant. For each sample, participants were presented with composite images generated by three models: GracoNet, CSANet, and our proposed method (GraPLUS) used identical foreground-background pairs, and were instructed to select the most realistic and reasonable composition. The final score for each model was calculated as the percentage of selections received across the entire set of 964 evaluations.

For generation diversity, we employ the Learned Perceptual Image Patch Similarity (LPIPS)~\cite{lpips}, computing the average perceptual distance between 10 different compositions generated per test sample by varying the random input vector. Higher LPIPS scores indicate greater diversity in the generated placements.

Spatial precision is evaluated through multiple geometric metrics that compared predicted versus ground-truth bounding boxes. We compute both mean Intersection over Union (IoU) and the percentage of predictions achieving IoU greater than 0.5. Center accuracy is measured via the mean Euclidean distance between the predicted and ground truth box centers (in pixels) and the percentage of predictions within 50 pixels of ground truth. Scale consistency is assessed through the ratio of smaller to larger box areas between predicted and ground-truth boxes (ensuring values between 0 and 1), alongside the percentage of predictions achieving a scale ratio greater than 0.8.

All evaluations use PyTorch's utilities, ensuring reproducible and consistent assessment of our model's performance across all critical aspects of the object placement task.

\subsection{Results and Analysis}

\begin{figure}
    \centering
    \includegraphics[width=0.95\textwidth]{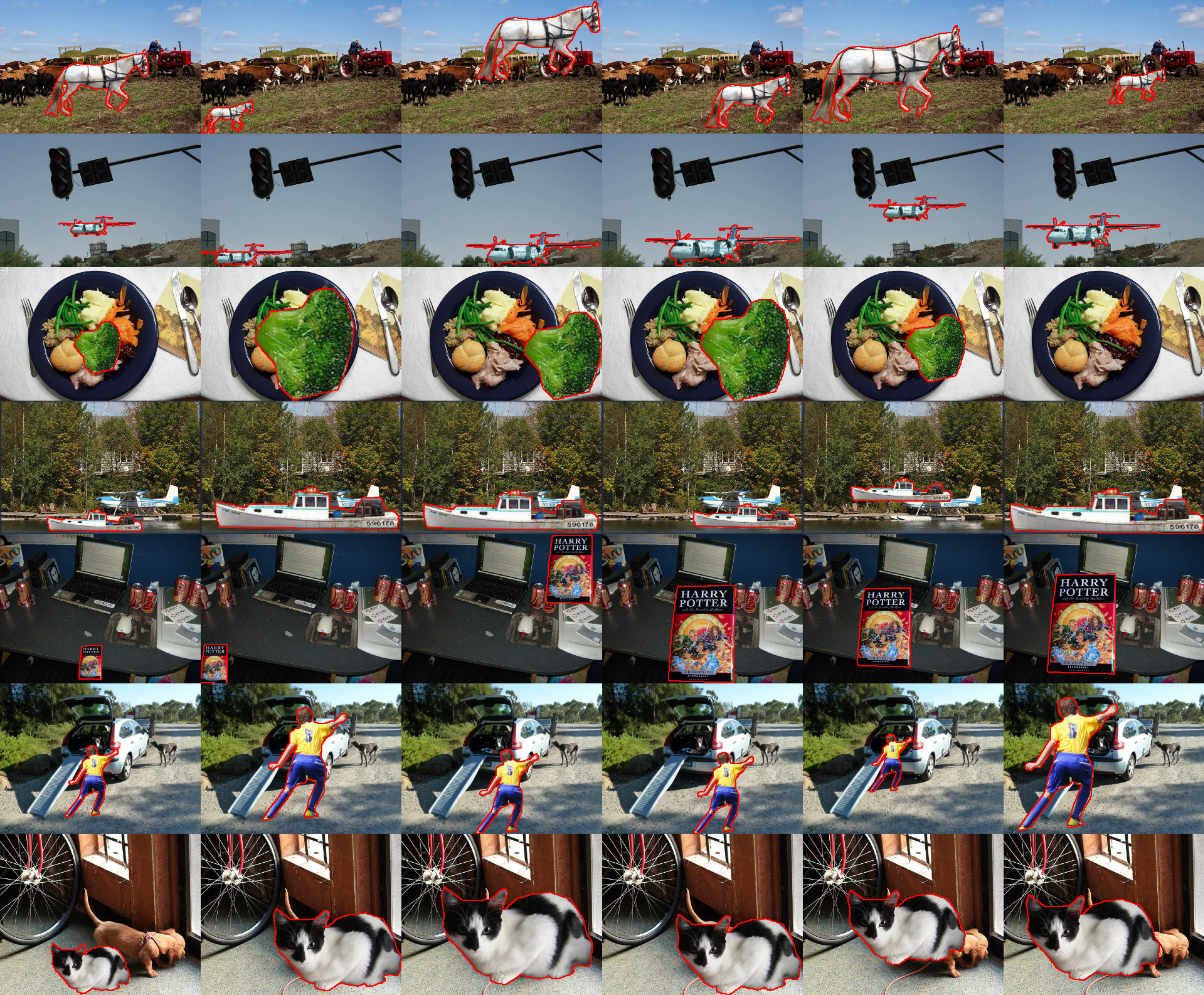}
    {\footnotesize
    \begin{tabular*}{0.95\textwidth}{@{}p{0.15\textwidth}p{0.114\textwidth}p{0.166\textwidth}p{0.095\textwidth}p{0.18\textwidth}p{0.085\textwidth}@{}}
        \centering GraPLUS & \centering CSANet & \centering CA\_GAN & \centering GracoNet & \centering PlaceNet & \centering TERSE
    \end{tabular*}
    }
    \caption{ Comparison of models for object placement. Each column corresponds to a model, and each row represents a specific index. The placed foreground objects are highlighted by the red outline, indicating the predicted placement boundaries. Best viewed with zoom-in.}
    \label{fig:model_comparisons}
\end{figure}

Our proposed approach demonstrates clear advantages in both semantic and spatial metrics when compared to prior works for which code is publicly available. Tables \ref{tab:generation_quality} and \ref{tab:spatial_precision} provide a comprehensive comparison with current models on various evaluation criteria. Furthermore, Figure~\ref{fig:model_comparisons} provides a visual comparison of object placements generated by different models, emphasizing the qualitative differences in placement plausibility and coherence.

\begin{table}
\renewcommand{\arraystretch}{1.2}
\caption{Comparison of models across generation quality metrics. Best results are \textbf{bolded}, second-best are \underline{underlined}.}
\label{tab:generation_quality}
\centering
\begin{tabular}{|l||c|c|c|c|}
\hline
\textbf{Model} & \textbf{User Study} $\uparrow$ & \textbf{Accuracy} $\uparrow$ & \textbf{FID} $\downarrow$ & \textbf{LPIPS} $\uparrow$ \\
\hline
TERSE & - & 0.683 & 47.44 & 0.000 \\
\hline
PlaceNet & - & 0.684 & 37.63 & 0.160 \\
\hline
GracoNet & \underline{0.263} & \underline{0.838} & 29.35 & 0.207 \\
\hline
CA-GAN & - & 0.734 & \underline{25.54} & \textbf{0.267} \\
\hline
CSANet & 0.216 & 0.803 & \textbf{22.42} & \underline{0.264} \\
\hline
\textbf{GraPLUS} & \textbf{0.521} & \textbf{0.921} & 28.83 & 0.055 \\
\hline
\end{tabular}
\end{table}

\begin{table}
\renewcommand{\arraystretch}{1.2}
\caption{Comparison of models across spatial precision metrics. Best results are \textbf{bolded}, second-best are \underline{underlined}.}
\label{tab:spatial_precision}
\centering
\begin{tabular}{|l||c|c|c|c|c|}
\hline
\textbf{Model} & \textbf{Mean IoU} $\uparrow$ & \textbf{IoU}$\geq$\textbf{0.5} $\uparrow$ & \textbf{C. Dist.} $\downarrow$ & \textbf{C. Dist.}$\leq$\textbf{50px} $\uparrow$ & \textbf{Scale}$\geq$\textbf{0.8} $\uparrow$ \\
\hline
TERSE & 0.171 & 10.9\% & 172.04 & 8.8\% & 9.1\% \\
\hline
PlaceNet & \underline{0.194} & 9.3\% & \underline{144.77} & \textbf{10.7\%} & 12.0\% \\
\hline
GracoNet & 0.192 & \underline{10.8\%} & 166.95 & 8.6\% & \underline{14.7\%} \\
\hline
CA-GAN & 0.165 & 9.2\% & 190.37 & 7.3\% & 12.5\% \\
\hline
CSANet & 0.162 & 9.4\% & 193.34 & 7.6\% & 13.6\% \\
\hline
\textbf{GraPLUS} & \textbf{0.203} & \textbf{11.7\%} & \textbf{141.77} & \underline{9.9\%} & \textbf{16.5\%} \\
\hline
\end{tabular}
\end{table}

In terms of placement quality, our model excels in generating semantically plausible object placements, outperforming previous methods with more consistent and realistic results. The results of the user study strongly support this finding, with GraPLUS receiving 52.1\% of the participant preferences, significantly outperforming GracoNet (26.3\%) and CSANet (21.6\%). This substantial preference margin confirms that human evaluators find our placements more natural and contextually appropriate. Although the visual quality, as indicated by the Fréchet Inception Distance (FID), shows competitive performance, our focus on semantic coherence maintains the balance between realism and visual appeal. The LPIPS score further supports this, reflecting our model's consistency in generating plausible compositions without sacrificing diversity.

Regarding spatial precision, our model achieves superior performance in both mean intersection-over-union (IoU) and center distance (C. Dist.) metrics, indicating a stronger ability to maintain correct spatial relationships between objects. This is particularly evident in our model's ability to preserve scale and maintain high-quality placements. Our model also demonstrates better adherence to scene boundaries, achieving exceptional alignment with object proportions and spatial positions, further solidifying its robustness in generating contextually appropriate placements.

In contrast, models like GracoNet and CSANet show strong visual quality but face limitations in maintaining semantic coherence or spatial precision, especially in terms of consistency and adherence to natural scene boundaries. PlaceNet, while competitive, still lags behind in generating more accurate spatial placements and producing semantically plausible compositions across different contexts.

These results underline the effectiveness of our semantic first approach, which leverages scene graph representations and cross-modal attention to optimize both object placement plausibility and spatial precision. By balancing these competing objectives, our model achieves a more holistic and robust generation of object placements compared to other methods.

\subsection{Ablation Studies}

We perform extensive ablation studies to validate our design choices and analyze the contribution of each component. Tables~\ref{tab:ablation_graph}--\ref{tab:ablation_arch} summarize our findings in different aspects of the model.

\begin{table}
\centering
\caption{Impact of scene graph node density on model performance.}
\label{tab:ablation_graph}
\begin{tabular}{lccc}
\hline
\textbf{Scene Graph Size} & \textbf{Acc (\(\uparrow\))} & \textbf{FID (\(\downarrow\))} & \textbf{LPIPS (\(\uparrow\))} \\
\hline
Sparse (10 Nodes) & 0.873 & 29.95 & \textbf{0.105} \\
Default (20 Nodes) & \textbf{0.921} & \textbf{28.83} & 0.055 \\
Dense (30 Nodes) & 0.865 & 31.71 & 0.055 \\
\hline
\end{tabular}
\end{table}

\textbf{Scene Graph Structure:} Table~\ref{tab:ablation_graph} demonstrates that a 20-node scene graph configuration optimally balances representation capacity and efficiency. Smaller graphs (10 nodes) lack sufficient representational power, while larger graphs (30 nodes) increase computational overhead without commensurate performance gains.

\begin{table}
\centering
\caption{Ablation study on Reconstruction Loss Weights.}
\label{tab:ablation_loss}
\begin{tabular}{lccc}
\hline
\textbf{$\lambda_{rec}$} & \textbf{Acc (\(\uparrow\))} & \textbf{FID (\(\downarrow\))} & \textbf{LPIPS (\(\uparrow\))} \\
\hline
0.0 & 0.866 & 34.08 & \textbf{0.131} \\
1.0 & 0.843 & 33.35 & 0.08 \\
10.0 & 0.859 & 31.79 & 0.071 \\
25.0 & 0.85 & 30.24 & 0.048 \\
50.0 (Default) & \textbf{0.921} & \textbf{28.83} & 0.055 \\
100.0 & 0.892 & 32.9 & 0.044 \\
\hline
\end{tabular}
\end{table}

\textbf{Loss Weighting:} As shown in Table~\ref{tab:ablation_loss}, the reconstruction loss weight $\lambda_{rec}$ significantly impacts model performance. The default value of 50.0 achieves optimal results with an accuracy of 92.1\% and a FID of 28.83, while lower or higher values lead to a decreased performance, particularly in the maintenance of diversity.

\begin{table}
\centering
\caption{Impact of Network Architecture Configuration on Model Performance}
\label{tab:ablation_attention}
\begin{tabular}{lccc}
\hline
\textbf{Network Configuration} & \textbf{Acc (\(\uparrow\))} & \textbf{FID (\(\downarrow\))} & \textbf{LPIPS (\(\uparrow\))} \\
\hline
\multicolumn{4}{l}{\textit{Graph Transformer Network (GTN) Layers}} \\
3 Layers & 0.815 & 29.77 & 0.053 \\
5 Layers (Default) & \textbf{0.921} & \textbf{28.83} & \textbf{0.055} \\
7 Layers & 0.797 & 30.19 & 0.041 \\
\hline
\multicolumn{4}{l}{\textit{GTN Attention Heads}} \\
4 Heads & 0.841 & 30.0 & \textbf{0.060} \\
8 Heads (Default) & \textbf{0.921} & \textbf{28.83} & 0.055 \\
12 Heads & 0.846 & 29.05 & 0.058 \\
16 Heads & 0.849 & 29.64 & 0.049 \\
\hline
\multicolumn{4}{l}{\textit{Multi-Head Attention (MHA) Heads}} \\
4 Heads & 0.866 & 29.76 & 0.055 \\
8 Heads (Default) & \textbf{0.921} & \textbf{28.83} & 0.055 \\
12 Heads & 0.846 & 30.02 & 0.05 \\
16 Heads & 0.838 & 39.98 & \textbf{0.06} \\
\hline
\end{tabular}
\end{table}

\textbf{Network Configuration:} Our experiments with GTN and MHA configurations (Table~\ref{tab:ablation_attention}) reveal that 5 GTN layers and 8 attention heads provide the best balance of performance and computational efficiency. Additional layers or heads yield diminishing returns while increasing model complexity.

\begin{table}
\centering
\caption{Ablation analysis of key model components.}
\label{tab:ablation_arch}
\begin{tabular}{lccc}
\hline
\textbf{Ablated Component} & \textbf{Acc (\(\uparrow\))} & \textbf{FID (\(\downarrow\))} & \textbf{LPIPS (\(\uparrow\))} \\
\hline
\multicolumn{4}{l}{\textit{Training Strategy}} \\
No Balanced Sampling & 0.873 & 29.14 & 0.053 \\
No Data Augmentation & 0.842 & 32.6 & 0.039 \\
\hline
\multicolumn{4}{l}{\textit{Feature Processing}} \\
No Spatial Features & 0.899 & \textbf{25.35} & 0.07 \\
Dynamic GPT-2 Embeddings & 0.883 & 25.67 & \textbf{0.078} \\ 
\hline
\multicolumn{4}{l}{\textit{Attention Architecture}} \\
No Position Encoding & 0.812 & 31.7 & 0.054 \\
No Residual Connection & 0.828 & 30.8 & 0.056 \\
\hline
\multicolumn{4}{l}{\textit{Discriminator Training}} \\
Single Real Update & 0.859 & 36.78 & 0.075 \\
\hline
Full Model & \textbf{0.921} & 28.83 & 0.055 \\
\hline
\end{tabular}
\end{table}

\textbf{Architecture Components:} From Table~\ref{tab:ablation_arch}, we observe that all the architecture elements contribute meaningfully towards the model's performance. The balanced sampling strategy is crucial, and its removal leads to a 4.8\% decrease in accuracy. Disabling data augmentation significantly affects model performance, reducing accuracy by 7.9\% and increasing FID to 32.6, while also decreasing generation diversity (LPIPS 0.039). Removing spatial information still maintains a reasonable accuracy (0.899), but freezing the GPT-2 initialized embedding layers results in a drop in accuracy to 0.883, suggesting that the pre-trained semantic representations from GPT-2 are more effective when preserved. The addition of positional encoding and residual connections in the MHA architecture is vital, as their removal causes large accuracy drops of 10.9\% and 9.3\%, respectively. Performing single discriminator updates on real samples, compared to double updates, results in lower accuracy (0.859) and much higher FID (36.78), but with an increase in generation diversity (0.075).

\section{Conclusion}

This work introduces a semantic-first framework for object placement that leverages scene graphs and language models to capture spatial and semantic relationships for contextually rich scene representation. Our results demonstrate significant improvements over existing approaches, achieving 92.1\% placement accuracy while maintaining competitive visual quality (FID: 28.83). The effectiveness stems from prioritizing semantic reasoning over pixel-level processing, though computational overhead from graph processing presents scalability challenges.

Future directions include multiple object placements, temporal consistency in videos, and few-shot learning for novel object categories, with promising applications in augmented reality, e-commerce, and computer vision data augmentation.

\bibliographystyle{plain}

\begin{thebibliography}{21}

\bibitem{terse}
S. Tripathi, S. Chandra, A. Agrawal, A. Tyagi, J. M. Rehg, and V. Chari,
\newblock Learning to generate synthetic data via compositing,
\newblock {\em arXiv preprint arXiv:1904.05475}, 2019.

\bibitem{placenet}
L. Zhang, T. Wen, J. Min, J. Wang, D. Han, and J. Shi,
\newblock Learning object placement by inpainting for compositional data augmentation,
\newblock {\em Computer Vision -- ECCV 2020}, A. Vedaldi, H. Bischof, T. Brox, and J.-M. Frahm, Eds. Cham: Springer International Publishing, 2020, pp. 566--581.

\bibitem{graconet}
S. Zhou, L. Liu, L. Niu, and L. Zhang,
\newblock Learning object placement via dual-path graph completion,
\newblock {\em European Conference on Computer Vision}. Springer, 2022, pp. 373--389.

\bibitem{ca_gan}
Y. Wang, Y. Feng, J. Wu, H. Xu, and J. Zheng,
\newblock CA-GAN: Object placement via coalescing attention based generative adversarial network,
\newblock {\em 2023 IEEE International Conference on Multimedia and Expo (ICME)}, 2023, pp. 2375--2380.

\bibitem{csanet}
Y. Wang, Y. Feng, and J. Zheng,
\newblock Learning object placement via convolution scoring attention,
\newblock {\em 35th British Machine Vision Conference, BMVC 2024, Glasgow, UK, November 25-28, 2024}. BMVA, 2024. [Online]. Available: https://papers.bmvc2024.org/0165.pdf

\bibitem{diffpop}
J. Liu, H. Zhou, S. Wei, and R. Ma,
\newblock DiffPop: Plausibility-guided object placement diffusion for image composition,
\newblock {\em arXiv preprint arXiv:2406.07852}, 2024. [Online]. Available: https://arxiv.org/abs/2406.07852

\bibitem{csenet}
Y. Qin, J. Xu, R. Wang, and X. Chen,
\newblock Think before placement: Common sense enhanced transformer for object placement,
\newblock {\em Computer Vision -- ECCV 2024}, A. Leonardis, E. Ricci, S. Roth, O. Russakovsky, T. Sattler, and G. Varol, Eds. Cham: Springer Nature Switzerland, 2025, pp. 35--50.

\bibitem{trad1}
T. Remez, J. Huang, and M. Brown,
\newblock Learning to segment via cut-and-paste,
\newblock {\em arXiv preprint arXiv:1803.06414}, 2018. [Online]. Available: https://arxiv.org/abs/1803.06414

\bibitem{trad2}
H. Wang, Q. Wang, F. Yang, W. Zhang, and W. Zuo,
\newblock Data augmentation for object detection via progressive and selective instance-switching,
\newblock {\em arXiv preprint arXiv:1906.00358}, 2019. [Online]. Available: https://arxiv.org/abs/1906.00358

\bibitem{trad3}
H.S. Fang, J. Sun, R. Wang, M. Gou, Y.L. Li, and C. Lu,
\newblock InstaBoost: Boosting instance segmentation via probability map guided copy-pasting,
\newblock {\em arXiv preprint arXiv:1908.07801}, 2019. [Online]. Available: https://arxiv.org/abs/1908.07801

\bibitem{trad4}
G. Georgakis, A. Mousavian, A. C. Berg, and J. Kosecka,
\newblock Synthesizing training data for object detection in indoor scenes,
\newblock {\em arXiv preprint arXiv:1702.07836}, 2017. [Online]. Available: https://arxiv.org/abs/1702.07836

\bibitem{trad5}
S.H. Zhang, Z. Zhou, B. Liu, X. Dong, D. Liang, P. Hall, and S.M. Hu,
\newblock What and where: A context-based recommendation system for object insertion,
\newblock {\em arXiv preprint arXiv:1811.09783}, 2018. [Online]. Available: https://arxiv.org/abs/1811.09783

\bibitem{topnet}
S. Zhu, Z. Lin, S. Cohen, J. Kuen, Z. Zhang, and C. Chen,
\newblock TopNet: Transformer-based object placement network for image compositing,
\newblock {\em arXiv preprint arXiv:2304.03372}, 2023. [Online]. Available: https://arxiv.org/abs/2304.03372

\bibitem{iopre}
S. Zhang, Q. Meng, Q. Liu, L. Nie, B. Zhong, X. Fan, and R. Ji,
\newblock Interactive object placement with reinforcement learning,
\newblock 2023.

\bibitem{opa}
L. Liu, Z. Liu, B. Zhang, J. Li, L. Niu, Q. Liu, and L. Zhang,
\newblock OPA: Object placement assessment dataset,
\newblock 2022. [Online]. Available: https://arxiv.org/abs/2107.01889


\bibitem{neural_motifs}
R. Zellers, M. Yatskar, S. Thomson, and Y. Choi,
\newblock Neural motifs: Scene graph parsing with global context,
\newblock {\em Proceedings of IEEE Conference on Computer Vision and Pattern Recognition}, 2018.

\bibitem{TDE}
K. Tang, Y. Niu, J. Huang, J. Shi, and H. Zhang,
\newblock Unbiased scene graph generation from biased training,
\newblock {\em CVPR}, vol. abs/2002.11949, 2020. [Online]. Available: https://arxiv.org/abs/2002.11949

\bibitem{radford2019language}
A. Radford, J. Wu, R. Child, D. Luan, D. Amodei, and I. Sutskever,
\newblock Language models are unsupervised multitask learners,
\newblock 2019. [Online]. Available: https://api.semanticscholar.org/CorpusID:160025533

\bibitem{stn}
M. Jaderberg, K. Simonyan, A. Zisserman, and K. Kavukcuoglu,
\newblock Spatial transformer networks,
\newblock 2016. [Online]. Available: https://arxiv.org/abs/1506.02025


\bibitem{fid}
M. Heusel, H. Ramsauer, T. Unterthiner, B. Nessler, and S. Hochreiter,
\newblock Gans trained by a two time-scale update rule converge to a local nash equilibrium,
\newblock 2018. [Online]. Available: https://arxiv.org/abs/1706.08500

\bibitem{lpips}
R. Zhang, P. Isola, A. A. Efros, E. Shechtman, and O. Wang,
\newblock The unreasonable effectiveness of deep features as a perceptual metric,
\newblock 2018. [Online]. Available: https://arxiv.org/abs/1801.03924

\end{thebibliography}


\end{document}